\pdfoutput=1

\documentclass[11pt]{article}

\usepackage{ACL2023}
\usepackage{times}
\usepackage{latexsym}

\usepackage{graphicx}
\graphicspath{ {./Figs/} }
\usepackage[T1]{fontenc}
\usepackage[utf8]{inputenc}
\usepackage{microtype}
\usepackage{inconsolata}

\usepackage{pifont}
\usepackage{latexsym}

\usepackage{balance}
\usepackage{helvet}  
\usepackage{courier}  
\usepackage{url}  
\usepackage{graphicx}  
\usepackage{amssymb}
\usepackage{amsmath}
\usepackage{algorithm}
\usepackage{algorithmic}
\usepackage{amsthm}
\usepackage{multirow}
\usepackage{pgffor}
\usepackage{graphicx}
\usepackage{epstopdf}
\usepackage{tikz}
\usepackage{epstopdf}
\usepackage{mathtools}
\usepackage{enumitem}
\usepackage{subfigure}
\usepackage{tabulary}
\usepackage{color}
\usepackage{url}
\usepackage{flushend}
\usepackage{stfloats}
\usepackage{tcolorbox}
\usepackage[utf8]{inputenc}
\usepackage[english]{babel}
\usepackage{tabularx}
\usepackage{CJKutf8}
\usepackage{microtype}
\usepackage{textcomp}
\usepackage{booktabs}
\usepackage{colortbl}
\usepackage{soul}
\usepackage{arydshln}
\usepackage{natbib}
\usepackage{appendix}

\usepackage{xspace}
\usepackage{microtype}
\usepackage{balance}
\usepackage{flushend}
\usepackage{mathrsfs}
\usepackage[utf8]{inputenc}

\newcommand{\ourapproach}{\textsc{TaG-QA}\xspace}
\newcommand{\cs}{\textsc{TaG-CS}\xspace}
\newcommand{\tableqa}{TableQA\xspace}
\newcommand{\fetaqa}{FeTaQA\xspace}

\newcount\DraftStatus  
\definecolor{darkgreen}{rgb}{0,0.5,0} 
\definecolor{purple}{rgb}{1,0,1} 
\definecolor{todocolor}{rgb}{0.9,0.1,0.1} 
\definecolor{fixcolor}{rgb}{0.1,0.7,0.3} 
\definecolor{wycolor}{rgb}{0.9,0.1,0.1} 
\definecolor{hycolor}{rgb}{0.7,0.7,0.3} 

\newcommand{\nbc}[3]{\ifnum\DraftStatus=1
	{\colorbox{#3}{\bfseries\sffamily\scriptsize\textcolor{white}{#1}}}
	{\textcolor{#3}{\sf\small$\blacktriangleright$\emph{#2}$\blacktriangleleft$}}
\fi}

\newcommand{\draftnote}[2]{\ifnum\DraftStatus=1
	\marginpar{
		\tiny\raggedright
		\hbadness=10000
		\def\baselinestretch{0.8}
		\textcolor{#1}{\textsf{\hspace{0pt}#2}}}
\fi}

\title{Localize, Retrieve and Fuse: A Generalized Framework for Free-Form Question Answering over Tables}

\author{
  Wenting Zhao$^1$~~~~ Ye Liu$^2$~~~~ Yao Wan$^3$~~~~ Yibo Wang$^1$ ~~~~ Zhongfen Deng$^1$~~~~ Philip S. Yu$^1$\\
  $^1$Department of Computer Science, University of Illinois at Chicago, IL, USA \\
  $^2$Salesforce Research, Palo Alto, USA \\
  $^3$School of Computer Sci. \& Tech., Huazhong University of Science and Technology, China\\
  \texttt{\{wzhao41,ywang633,zdeng21,psyu\}@uic.edu} \\
  \texttt{yeliu@salesforce.com}, \texttt{wanyao@hust.edu.cn}
}

\begin{document}
\maketitle
\begin{abstract}

Question answering on tabular data (\textit{a.k.a} \tableqa), which aims at generating answers to questions grounded on a provided table, has gained significant attention recently.
Prior work primarily produces concise factual responses through information extraction from individual or limited table cells, lacking the ability to reason across diverse table cells. Yet, the realm of free-form \tableqa, which demands intricate strategies for selecting relevant table cells and the sophisticated integration and inference of discrete data fragments, remains mostly unexplored.
To this end, this paper proposes a generalized three-stage approach: \textit{\underline{Ta}ble-to-\underline{G}raph conversion and cell localizing, external knowledge retrieval, and the fusion of table and text (called \ourapproach)}, to address the challenge of inferring long free-form answers in generative \tableqa. 
In particular, \ourapproach (1) locates relevant table cells using a graph neural network to gather intersecting cells between relevant rows and columns, (2) leverages external knowledge from Wikipedia,
and (3) generates answers by integrating both tabular data and natural linguistic information.
Experiments showcase the superior capabilities of \ourapproach in generating sentences that are both faithful and coherent, particularly when compared to several state-of-the-art baselines. Notably, \ourapproach surpasses the robust pipeline-based baseline TAPAS by 17\% and 14\% in terms of BLEU-4 and PARENT F-score, respectively. Furthermore, \ourapproach outperforms the end-to-end model T5 by 16\% and 12\% on BLEU-4 and PARENT F-score, respectively.\footnote{Source code will be released at \url{https://github.com/wentinghome/TAGQA}.}


\end{abstract}

\section{Introduction}

Question answering is to generate precise answers by interacting efficiently with unstructured, structured, or heterogeneous contexts, such as paragraphs, knowledge bases, tables, images, and various combinations thereof
\citep{burke1997question,yao-van-durme-2014-information,talmor2021multimodalqa,hao-etal-2017-end}. 
Among these, question answering on tabular data (\tableqa) is a challenging task that requires the understanding of table semantics, as well as the ability to reason and infer over relevant table cells~\citep{herzig-etal-2021-open,chen2020hybridqa, chen-etal-2021-finqa}.

For the task of \tableqa, from our investigation, most current studies are focusing on the factoid \tableqa, in which the answer is in a few words or a phrase copied directly from relevant table cells.
In particular, current works on factoid \tableqa are mainly categorized into two groups: (1) pipeline-based methods consisting of two stages, i.e., cell retrieval and answer reader~\citep{zhu-etal-2021-tat,chen2020open}; and (2) end-to-end neural networks such as a paradigm of sequence-to-sequence model that takes the context of question answering (e.g., question and table cells) as input to generate natural-language answers~\citep{li2021tsqa,pan2022end,herzig-etal-2021-open, pan-etal-2021-cltr,chen-2023-large}.

\begin{figure}[t]
    \centering
    \includegraphics[width=0.48\textwidth]{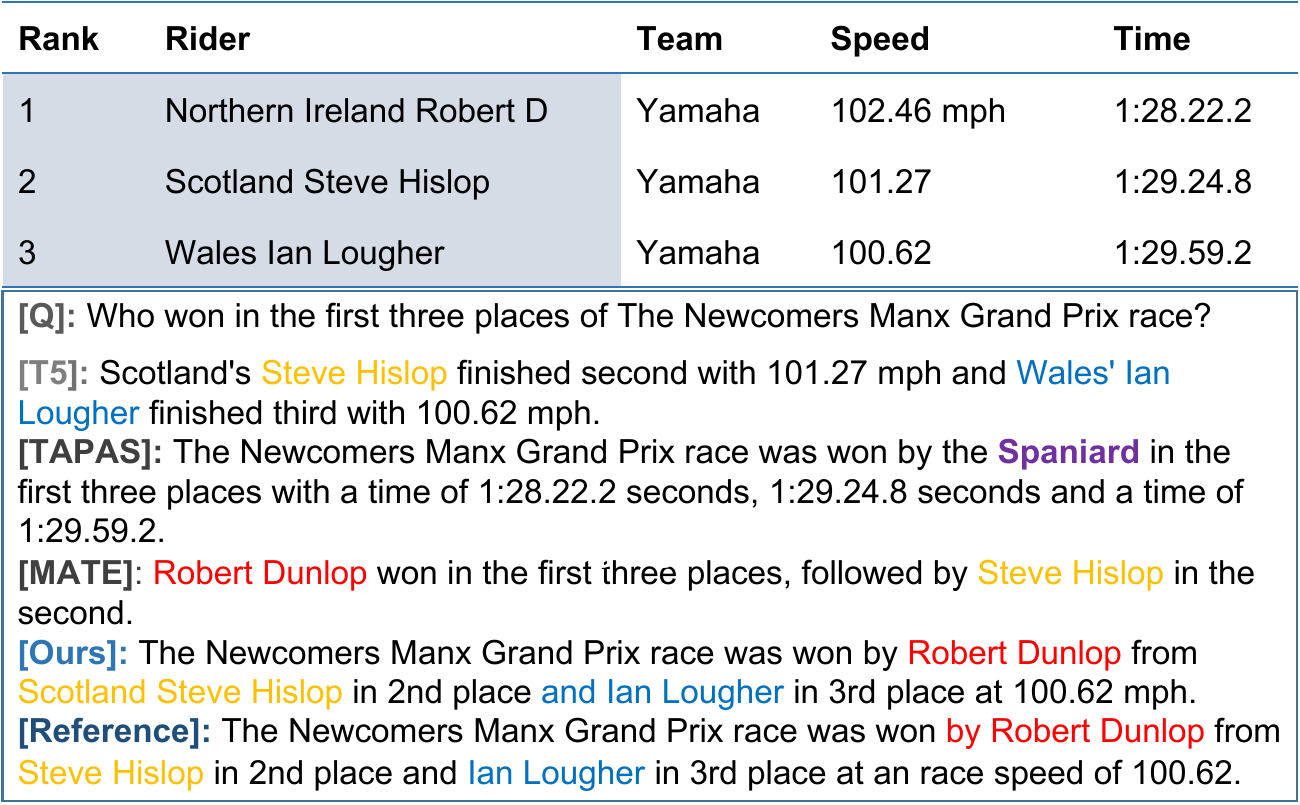}
    \caption{A motivating example to show the insights of our proposed approach when comparing with several state-of-the-art methods.}
    \label{fig:intro_1}
\end{figure}
 
 
Despite much progress made on factoid TableQA, a contradiction between the factoid TableQA and TableQA exists in real scenarios.
In factoid TableQA, the answers are always in a short form with a few words directly copied from the relevant table cells.
However, in real-world scenarios, 
the answers are expected to be long and informative sentences in a free form, motivating us to target the free-form TableQA in this paper.

It is challenging to generate coherent and faithful free-form answers over tables.
(1) \textit{The well-preserved spatial structure of tables is critical for retrieving relevant table cells to the question.} 
Different from factoid \tableqa, free-form \tableqa with sophisticated question shares less semantic similarities to the table content, while depending more on the spatial structure of tables to infer multiple related cells such that the related cells may be located in a relatively connected area, e.g., from either a few selected rows or columns.
(2) \textit{The selected table cells, containing the key point, are insufficient for composing the entire coherent sentences.} To generate fluent natural-language sentences as answers, external information such as the relevant background knowledge about the question is necessary.
(3) \textit{It is expected to aggregate and reason from the question, retrieved table cells, and external knowledge to compose a reasonable answer.} Given the heterogeneous information, a practical model should be capable of aggregating the information efficiently and generating a coherent and fluent free-form answer.
\begin{figure*}[t]
\centering
\includegraphics[width=0.98\textwidth]{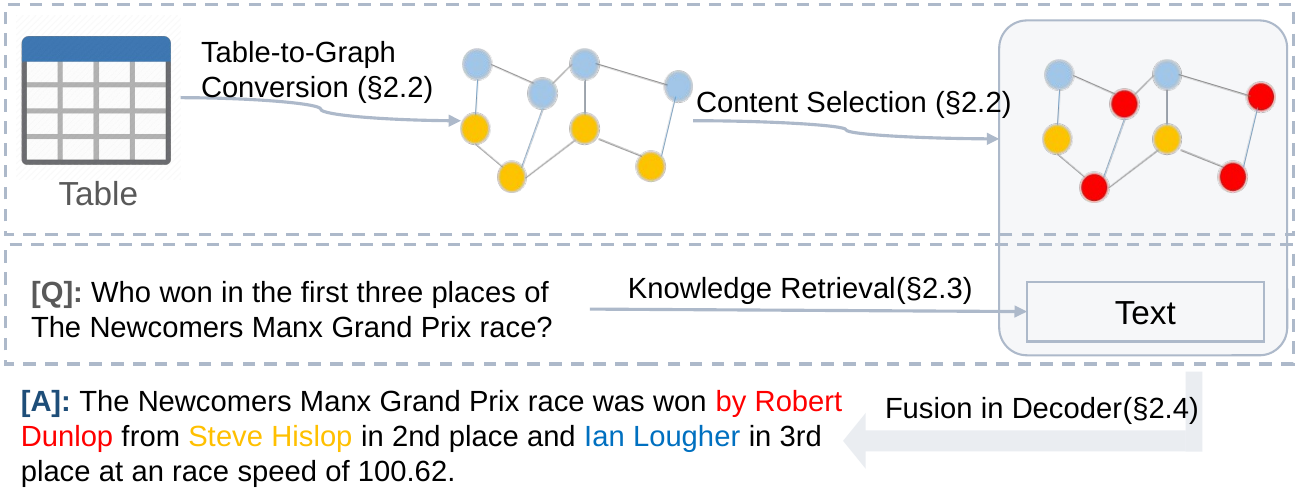}
\caption{An overview of \ourapproach. The input to \ourapproach is a combination of one table and question, while the output is an answer. The top box shows the content selection process which first converts the table to a graph and selects relevant nodes using GNN. The middle box shows the process of using the spare retrieval technique to retrieve relevant text as complementary information. The rightmost blue box is to integrate the selected cells and retrieved texts to generate the final answer.}
\label{fig:model_overview}
\centering
\end{figure*}

Figure~\ref{fig:intro_1} provides a motivating example to illustrate the insights of this paper.
Given a table describing ``\textit{the 1983 Manx Grand Prix Newcomers Junior Race Results}'' and a question ``\textit{Who won in the first three places of The Newcomers Manx Grand Prix race?}'', the goal is to select relevant cells first and then generate a natural sentence as an answer. 
From this table, we can observe that the state-of-the-art model TAPAS and MATE only select the ``\textit{rider}'' while missing the ``\textit{rank}'' column, providing low cell selection coverage. For the overall generation quality, we can observe that both the end-to-end T5~\citep{t5} and the pipeline-based TAPAS~\citep{tapas} and MATE~\citep{mate} are missing key information from the table by merely mentioning part of the three riders. 
In addition, the TAPAS introduces a hallucinated rider named ``\textit{Spaniard}''. These observations motivate us to design a model that can select the relevant cells more accurately and generate faithful answers grounded on the table given a question.

Based on the aforementioned insights, this paper designs a three-stage pipeline framework to tackle the problem of free-form \tableqa. Even though the end-to-end \tableqa models with high accuracy are prevalently ascribed to the suppression of error accumulated from one-stage training, the long table distracts the model from focusing on relevant table cells, resulting in irrelevant answers. On the other hand, the cell selection module provides a controllable and explainable perspective by extracting a small number of table cells as anchors for the model to generate answers. 
For the content selection stage, inspired by the recent success of graph models, we convert the table to a graph by designing the node linking and applying a Graph Neural Network (GNN) to aggregate node information and classify whether the table cell is relevant or not. 
In addition, to generate informative free-form answers, we employ a spare retrieval technique to explore extra knowledge from Wikipedia. Consequently, both the extra knowledge and relevant cells are taken into account to calibrate the pre-trained language model bias. Lastly, we adopt a fusion layer in the decoder to generate the final answer.

To summarize, the primary contributions of this paper are three-fold. 
(1) To the best of our knowledge, we are the first to convert a semi-structured table into a graph, and then design a graph neural network to retrieve relevant table cells. 
(2) External knowledge is leveraged to fill in the gap between the selected table cell and the long informative answer by providing background information. 
(3) Comprehensive experiments on a public dataset named FeTaQA~\citep{fetaqa} are performed to verify the effectiveness of \ourapproach. Experimental results show that \ourapproach outperforms the strong baseline TAPAS by $17\%$ and $14\%$, and outperforms the end-to-end T5 model by $16\%$ and $12\%$, in terms of BLEU-4 and PARENT F-score, respectively.

\section{\ourapproach Approach}
In this section, we first formulate the problem of TableQA, and introduce the details of our proposed approach \ourapproach.
\subsection{Problem Formulation}
A free-form question-answering task is formulated as generating an answer $a$ to a question $q$ based on a semi-structured table $T$ including table cell content and table meta information such as column, and row header. Different from the factoid table question answering task with a short answer, the free-form QA aims at generating informative and long answers.

\subsection{Overview}
Figure~\ref{fig:model_overview} illustrates the overall architecture of our proposed \ourapproach, which is composed of three stages, i.e., relevant table cell localization, relevant external knowledge retrieval, and table-text fusion. 
\textit{(1) Relevant table cell localization.} We first propose a table-to-graph converter to transform a table into a graph which can preserve the table's spatial information. We think that the graph-based table representation can better assist in selecting relevant table cells.
\textit{(2) External knowledge retrieval.} We adopt the sparse retrieval technique to collect external information which can be complementary information for the final answer generation. 
\textit{(3) Table-text fusion.} We employ the fusion-in-decoder model by taking both the selected table cells and the external sources into account to generate the answer. 
The above three steps enable our model to generate a faithful free-form answer for a question grounded on the table.

\subsection{Relevant Table Cell Localization}
The initial phase of \ourapproach involves table content selection, a pivotal step that serves as the foundation for subsequent stages. Notably, this stage is of utmost importance as it supplies essential input to the subsequent processes. \fetaqa presents a formidable challenge as a dataset, with a Median/Avg percentage of relevant table cells at $10.7\% / 16.2\%$. In order to enhance the precision of the content selection stage, we design a table-to-graph converter to preserve the inherent spatial structure of the tables. We employ GNN to effectively aggregate information at the cell level and subsequently perform a classification task on the table cells.

\paragraph{Table-to-Graph Converter}\label{sec:Table-to-Graph Converter}
\begin{figure}[!t]
    \centering
    \includegraphics[width=0.5\textwidth]{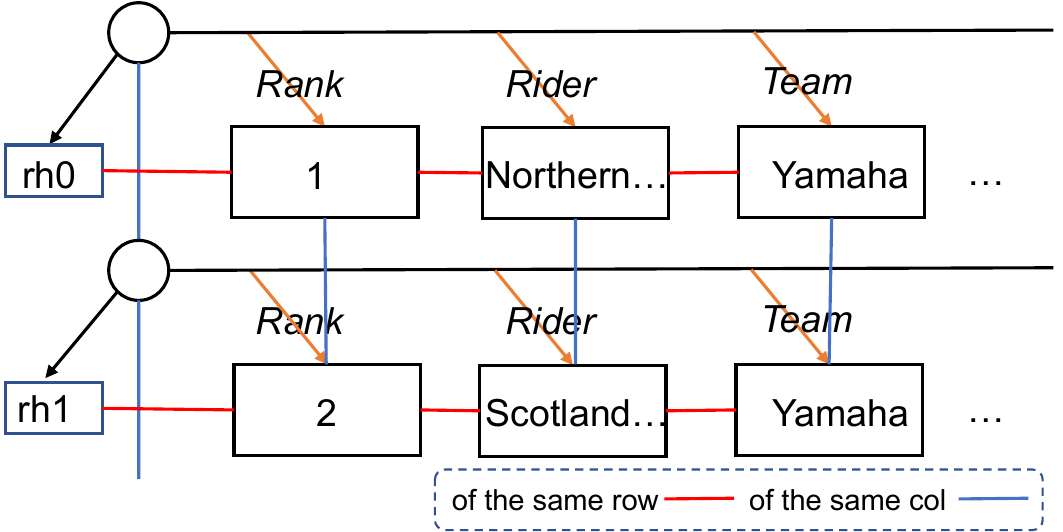}
    \caption{
Convert the table shown in Figure \ref{fig:intro_1} into a graphical representation. 
    ``\textit{rh0}'' is the added row header for the first row. Two relations are considered in the table graph, i.e., ``\textit{of the same row}'' and ``\textit{of the same column}'' relations.}
    \label{fig:method_table_transformation}
\end{figure}

State-of-the-art models prefer to adopt the pre-trained Language Models (LMs) to make predictions by transforming the semi-structured table into natural sentences using a pre-defined template
. However, they lose the table structure information and deteriorate the performance of downstream tasks. 

\ourapproach designs a table-to-graph converter to transform a table into a graph, preserving the table structure by identifying the cell-to-cell relations.
Figure~\ref{fig:method_table_transformation} shows an example of transforming a table into a graph. For the $i$-th row, we add an empty row header as $rhi$ which reflects the entire row information. All the table cells from the same row are fully connected, and all the table cells from the same column are also fully connected. Besides, we design two types of relations for the table graph, i.e., ``\textit{of the same row}'' and ``\textit{of the same column}'' relations. In particular, ``\textit{of the same row}'' relation captures the entity information, while ``\textit{of the same column}'' relation reveals the connection of the same attribute. 

In addition, to incorporate the question node into the graph, we create a question node and assign a linking edge between the question and each table cell with the relation ``\textit{question to cell}''.

\paragraph{\ourapproach Content Selection}
Inspired by QA-GNN \citep{yasunaga-etal-2021-qa}, we propose a content selection module (\cs) that retrieves relevant table cells from the table-based graph. \cs takes the converted table graph from Sec.~\ref{sec:Table-to-Graph Converter} as input, and outputs the question-related table cells. \cs reasons over the table cell level, and each graph node represents a table cell. To fully explore the table semantic and the spatial information, \cs acquires the initial graph node embedding through a pre-trained LM e.g., BERT. Besides, the pre-trained LM and GNN are jointly trained to predict the selected cells. 

\paragraph{GNN Architecture}
We use Graph Attention Network (GAT)~\citep{velivckovic2017graph} which leverages masked self-attention layers and employs iterative message passing among neighbors is applied to predict the selected graph node. GAT follows Eq.~\ref{eq:tablegnn-main} to update the $i$-th node feature $h_i^l \in \mathbb{R}^{D}$ at layer $l$ through gathering the weighted attention among its neighbors $\mathcal{N}_i$.
\begin{equation}
\label{eq:tablegnn-main}
h_i^l=f_g\left(\sum_{s\epsilon N_t \cup \{t\}}\alpha_{st}m_{st}\right)+h_t^{l-1}
\end{equation}
where $\alpha_{st}$ and $m_{st}\in \mathbb{R}^{N}$ are the self-attention weight and the message passed from source node $s$ to target node $t$ respectively, and $f_g$ is a 2-layer Multi-Layer Perceptron (MLP) with batch normalization.
The message $m_{st}\in \mathbb{R}^{N}$ from node $v_s$ to $v_t$ is computed using Eq.~\ref{eq:tablegnn-message}. 
\begin{equation}
\label{eq:tablegnn-message}
m_{st}=f_m(h_s^{l-1}, u_s, r_{st})
\end{equation}
where $u_s\in \mathbb{R}^{T/2}$ is the source node $s$ feature linearly transformed from the one hot vector node type $u_t$. $r_{st}\in \mathbb{R}^{T}$ is the relation feature from source node $s$ to target node $t$ computed through a 2-layer MLP by taking relation type, source, and target node type into account. $f_m$ is a linear transformation.

The self-attention coefficient $\alpha_{st}$ is updated in Eq.~\ref{eq:tablegnn-alpha}. Query and key vectors are linearly transformed by $g_q$ and $g_k$, as node, edge feature, and the previous layer hidden state provided.
\begin{equation}
\label{eq:tablegnn-alpha}
\alpha_{st}=\frac{exp(\gamma_{st})}{\sum_{{t}'\epsilon N_s\cup  \{s\}}exp({\gamma_{st}}')) },  \gamma_{st} = \frac{Q_s^TK_t}{\sqrt{N}}
\end{equation}
\begin{equation}
\label{eq:tablegnn-q}
Q_{s}=g_q(h_s^{l-1}, u_s, r_{st})
\end{equation}
\begin{equation}
\label{eq:tablegnn-k}
K_{t}=g_k(h_s^{l-1}, u_t, r_{st})
\end{equation}


\paragraph{GNN Training and Inference}
Given a question $q$ and a table $T$, \cs reasons over a graph containing both the table cell nodes and the question node by making predictions on the row and column level. We observe that relevant table cells tend to show up in a relatively connected area, thus we make predictions over row and column headers and choose the intersection area. Compared to predicting over the cell level which results in low recall, our method gains a higher chance to capture relevant table cells. For the training stage, \cs maximizes the cross entropy to predict the row and column for relevant cells.

\subsection{External Knowledge Retrieval}
\ourapproach is the first attempt to leverage the external knowledge to address the table-based free-form QA task. \ourapproach adopts an effective and simple Spare Retrieval based on the TF/IDF approach to select a potentially relevant context from Wikipedia.

\paragraph{Sparse Retrieval}
For \ourapproach, the external knowledge is served as a complimentary background context for the next table and text fusion stage. We choose the spare retrieval method using BM25~\citep{robertson2009probabilistic} as a ranking function to retrieve the most relevant text as supplementary information. Given a query $q$ with $m$ keywords $k_1, k_2, \ldots , k_m$, the BM25 ranking score $p_i$ for document $d_i$ is calculated by Eq.~\ref{eq:bm25}, 
\begin{equation}
\label{eq:bm25}
p_i=\sum_{j=1}^{m}\frac{idf(q_j)\times tf(q_j,d_i)\times (\alpha+1)}{tf(k_j,d_i)+\alpha (1-\beta+\beta\frac{|d_i|}{L_D})}
\end{equation}
where $idf$ is the Inverse Document Frequency~(IDF), $tf(k_j,d_i)$ is the term frequency of the keyword $k_j$ in document $d_i$, and $L_D$ is the average document length. 

\subsection{Table-Text Fusion}
After obtaining the predicted highlighted table cells from the table as well as the support context from Wikipedia, \ourapproach aggregates and combines the two information sources through a sequence-to-sequence model Fusion-in-Decoder (FiD) \citep{izacard-grave-2021-leveraging}. 
FiD appends the question to each information source, encoding each component independently. It subsequently merges all source features and transmits them to the decoder.

\paragraph{Fusion in Decoder}
Fusion-in-Decoder based on T5~\citep{t5} architecture takes question, support context, and the retrieved semi-structured table cells as input. We flatten the highlighted cells as a natural sentence to fit with its pre-trained LM architecture.
For the table example shown in Figure~\ref{fig:intro_1}, the ground-truth selected cells from the first two columns ``\textit{Rank}'' and ``\textit{Rider}'' can be linearized as ``\textit{Rank is 1 \texttt{[SEP]} Rider is Northern Ireland Robert D \texttt{[SEP]} Rank is 2 \texttt{[SEP]} Rider is Scotland Steve Hislop \texttt{[SEP]} Rank is 3 \texttt{[SEP]} Rider is Wales Ian Loug.}'', where \texttt{[SEP]} is a special token to indicate the end of table slot value.

\begin{table}
\centering
\resizebox{0.98\textwidth/2}{!}{
\begin{tabular}{lccc}
\bottomrule[1.5pt]
& \textbf{Precision} & \textbf{Recall} & \textbf{F-1}\\
\hline
TAPAS~\citep{tapas} & \textbf{65.31}  & 24.20  &  35.32\\ 
MATE~\citep{mate} &  56.93 & 22.21  &  31.95 \\ 
\ourapproach(Ours) &  47.60 & \textbf{43.06}  &  \textbf{45.22} \\ 

\bottomrule[1.5pt]
\end{tabular}
}
\caption{\label{tab:cs-acc}
Content selection results on \fetaqa dataset. 
}
\end{table}
\begin{table*}
\centering
\setlength{\tabcolsep}{4pt} 
\resizebox{0.99\textwidth}{!}{
\begin{tabular}{lllllll}
\bottomrule[1.5pt]
  &  & \textbf{BLEU-4} & \textbf{METEOR} & \textbf{ROUGE-L} & \textbf{PARENT (P/R/F)} & \textbf{PARENT-T (P/R/F)}\\
\hline
& \textbf{\textit{End-to-end}}  &   &      &   &     &   \\
\hline
\multirow{3}{*}{UniLM} & Q-fullTab & 17.57  & 28.30  & 39.46  & 38.21/24.18/25.56  &  26.48/53.99/33.70  \\ 
   &  Q-Retrieve  &  18.46 &  27.21 & 39.36 &  34.12/23.42/23.76 &  20.37/43.41/25.69  \\ 
 & Q-Retrieve-fullTab & 18.89  & 26.86  & 38.86  &  35.29/23.17/23.72 &  22.07/44.83/27.44    \\ 
 \hline

\multirow{3}{*}{BART} & Q-fullTab  & 7.62  & 25.70  &  25.76  &  39.64/19.68/22.62  &   25.77/39.53/28.78\\ 
   &  Q-Retrieve &  12.20 & 25.15  & 28.27 & 35.55/20.67/22.37 &  18.31/31.07/20.94  \\ 
 & Q-Retrieve-fullTab & 11.97  & 26.41  &  28.24 &38.45/22.12/23.96  & 20.57/34.36/23.46\\ 

 \hline

\multirow{3}{*}{T5} & \textbf{Q-fullTab}*  &  15.66 & 21.80 &  35.48  & 38.88/14.83/18.01 & 25.11/33.62/26.17 \\
   &  Q-Retrieve  &  25.17 & 24.87  & 39.89 & 33.54/20.3/21.68 &  17.35/31.21/20.13  \\ 
 & Q-Retrieve-fullTab & \underline{27.60}  & \underline{26.71} & \underline{42.38}  & \underline{38.49/23.2/25.06} & \underline{20.98/35.79/24.02}  \\ 

\hline

\multirow{2}{*}{\textbf{Oracle-T5}}& Q-OracleCell &  21.77 & 28.35  & 42.54  &   53.37/26.39/30.61   & 38.21/54.22/41.49   \\
 &  Q-Retrieve  &  25.17 & 24.87  & 39.89 & 33.54/20.3/21.68 &  17.35/31.21/20.13  \\ 
 & Q-Retrieve-OracleCell&  31.00  & 30.35  & 46.72  & 46.3/28.44/30.93 &  27.07/44.32/30.71  \\ 

\hline
& \textbf{\textit{Pipeline}}  &   &      &   &     &   \\
\hline
\multirow{2}{*}{TAPAS-T5} & Q-predCell &   14.50 & 21.18  & 35.51 &  39.14/12.34/15.67   & 25.19/29.47/24.38  \\
& Q-Retrieve-predCell &  26.81  & 26.92  & 42.59 & 39.23/21.96/24.15    &  21.43/34.54/23.61 \\
\hline

\multirow{2}{*}{MATE-T5} & Q-predCell &  14.28   & 21.01  & 35.36 &    39.07/12.2/15.53 &  24.83/29.56/24.25 \\
 & Q-Retrieve-predCell &  26.85    &  26.96 & 42.60 &  39.05/21.89/23.99   & 21.1/34.62/23.57  \\
\hline

\multirow{2}{*}{TAGQA-T5} & Q-predCell   &   17.08  & 23.22  & 38.38 & 41.84/16.53/20.1    & 27.11/37.03/28.45  \\
 & Q-Retrieve-predCell    &  \underline{28.01}($\uparrow$ 0.41)    &   \underline{27.91}($\uparrow$ 1.20) & \underline{44.16}($\uparrow$ 1.78) &  \underline{41.35/23.87/26.2}($\uparrow$ 1.14)   &   \underline{22.89/37.29/25.64}($\uparrow$ 1.64) \\
 
\hline
TAGQA-FiD & \textbf{Q-Retrieve-predCell*}   & \textbf{31.84}($\uparrow$ 16.18)  & \textbf{30.16}($\uparrow$ 8.36)  &  \textbf{49.39}($\uparrow$ 13.91)   & \textbf{47.56}/\textbf{26.20}/\textbf{29.59}($\uparrow$ 11.58) &   \textbf{25.44}/\textbf{39.11}/\textbf{28.26}($\uparrow$ 2.09)  \\

\bottomrule[1.5pt]
\end{tabular}

}

\caption{\label{tab:res-main}
Results on \fetaqa dataset. ``P/R/F'' denotes the precision/recall/F score. We report end-to-end model UniLM, BART and T5, and the pipeline results. The results of various table cell selection strategies TAPAS, MATE and our proposed TAG with T5 as backbone generation model are noted as TAPAS-T5, MATE-T5 and TagQA-T5. To validate the effectiveness of proposed framework components, we test different combinations of source information to models where ``Q'' is question, ``Retrieve'' is the retrieved external knowledge, ``fullTab'' is full table, and ``predCell'' refers to the selected table cell. And the last row TAGQA-FiD is the proposed method.
}
\end{table*}
\begin{table}
\centering
\resizebox{0.98\textwidth/2}{!}{
\begin{tabular}{lccccc} 
\bottomrule[1.5pt]
Method  & Overall \\
\hline
Reference &  4.94   \\
\hline
UniLM [end-to-end]~\citep{unilm} & 3.88  \\
BART [end-to-end]~\citep{lewis-etal-2020-bart}  & 3.67 \\ 
T5 [end-to-end]~\citep{t5} & 3.81 \\ 
\hline
Tapas [pipeline]~\citep{tapas} &  3.38  \\
MATE [pipeline]~\citep{mate}  & 3.30  \\
\ourapproach [pipeline] &  \textbf{3.93}   \\
\bottomrule[1.5pt]
\end{tabular}
}
\caption{\label{tab:human-evaluation}
Results of human evaluation for reference, end-to-end model and pipeline methods. \ourapproach outperforms the pipeline models by a large margin, and achieves performance on par with the strong end-to-end baseline model T5.
}
\end{table}

\section{Experiments and Analysis}

In this section, we explore the following experimental questions: (1) Does proposed \ourapproach generate a more coherent and faithful answer compared with the baseline? (2) Is table cell selection, knowledge retrieval, and fusion necessary for the free-form TableQA? (3) Is it promising to keep enhancing the three modules of \ourapproach?

\subsection{Dataset}

This paper focuses on tackling the challenge of generating long free-form answers, rather than the short factoid responses. Consequently, we have opted for the utilization of the state-of-the-art dataset, \textit{FetaQA}~\citep{fetaqa}, as our testbed. The training dataset comprises 7,327 instances, while the development and test sets encompass 1,002 and 2,004 examples, respectively.

\subsection{Implementation Details}
\noindent \emph{\textbf{\cs)}} 
\cs applies BERT checkpoint ``\texttt{bert-based-uncased}'' to learn the table cell representation. For the BERT model, we set the learning rate to $1e\text{-}6$ and impose a maximum token length of 35 for each cell. Subsequently, the acquired table cell-level embeddings serve as input node features for our GNN. Within the \cs framework, our GNN module comprises 3 layers, each with node features of 200 dimensions. Additionally, we apply a dropout rate of 0.2 to each layer for regularization.

We train our model on the \fetaqa dataset, configuring it to run for a maximum of 50 epochs. We employ the RAdam optimizer~\citep{liu2019variance} with a weight decay of 0.01, utilizing a powerful 24G memory Titan-RTX GPU.
To optimize GPU memory usage, we set the maximum number of table cells as 200 and set the batch size as 1. 
The selection of the best checkpoint is based on the performance of the model on the development set, which is then used for decoding the test set.
Additionally, to enhance efficiency, \cs is employed to select intersection cells from the top 3 rows and 3 columns as the relevant cells, drawing upon our accumulated experience in this context.

\noindent \emph{\textbf{Sparse Retrieval)}} 
Our implementation relies on the PyTorch-based toolkit Pyserini, designed for reproducible information retrieval research using both sparse and dense representations. We utilize the question as the query to retrieve pertinent contextual information from Wikipedia, selecting the first sentence from the top results. We specifically employ the Lucene Indexes, denoted as ``enwiki-paragraphs''\footnote{https://github.com/castorini/pyserini}.

\noindent \emph{\textbf{FiD)}} 
In the context of FiD, \ourapproach employs the Adam optimizer with a learning rate of $1e\text{-}5$. We select the best checkpoint for inference purposes. In the inference phase, we utilize beam search with a beam size of 3 and apply a length penalty of 1 when generating answers.

\subsection{Baselines}

To validate the effectiveness of \ourapproach, we choose two different types of methods as baselines, including end-to-end and pipeline-based models. 

Firstly, we compare \ourapproach with strong state-of-the-art end-to-end pre-trained generative LMs.  
UniLM~\citep{unilm}, BART~\citep{lewis-etal-2020-bart}, and T5 ~\citep{radford2019language}.
For the input format to the end-to-end model, we flatten the table by concatenating special token \texttt{[SEP]} in between different table cells, and concatenate with the question as a natural sentence, e.g. ``\textit{question \texttt{[SEP]} flattened table}''.
Furthermore, we compare the performance of our proposed model with pipeline-based methods which include two stages: content selection and answer generation. Content selection makes predictions of relevant cells. We choose two table-based pre-training models: TAPAS~\citep{tapas} and MATE~\citep{mate}.
Moreover, T5 is chosen as the baseline model's answer generation backbone due to the integration capacity for the table cell and retrieved knowledge.

\subsection{Automatic Evaluation Metrics}

We use various automatic metrics to evaluate the model performance. Due to the pipeline style of \ourapproach, we report two sets of metrics for content selection and answer generation stages respectively.
Firstly, to evaluate the retrieval competency of the table semantic parser, we report Precision, Recall, and F1 scores. Besides, to evaluate the answer generation quality, we choose several automatic evaluation metrics, i.e., BLEU-4~\citep{papineni2002bleu}, ROUGE-L~\citep{lin2004rouge} and METEOR~\citep{banerjee2005meteor}, to evaluate the n-gram match between the generated sentence and the reference answer. Considering the limitation that those metric fails to reflect the faithfulness answer to the fact from the table, we report PARENT~\citep{dhingra-etal-2019-parent} and PARENT-T~\citep{wang-etal-2020-towards} score. PARENT score takes the answer matching with both the reference answer and the table information into account, while PARENT-T focuses on the overlap between the generated answer with the corresponding table.

\subsection{Results}
 We first evaluate the \cs content selection stage table semantic parsing results, as shown in Table~\ref{tab:cs-acc}. For the F-1 score, \ourapproach outperforms the strong baseline model TAPAS and MATE by 9.9\% and 13.27\%. For recall, TAG-QA achieves the best result, demonstrating that \ourapproach retrieves more relevant table cells. For precision, the baseline model outperforms \ourapproach by retrieving fewer cells which includes more relevant cells. However, the low precision and high recall are a trade-off since the relevant cells make a stronger impact on the overall answer generation quality. Thus, we can tolerate a small amount of irrelevant cells and keep the correct cells as many as possible.

In addition, Table~\ref{tab:res-main} shows the measurements of generated answer quality using \ourapproach compared to previous both end-to-end and pipeline-based state-of-the-art models. 
From overlapping-based metrics BLEU-4, METEOR, and ROUGE-L, \ourapproach outperforms all the end-to-end and pipeline-based models. Specifically, \ourapproach gains 14.27\%/1.86\%/9.93\% more than the best end-to-end model UniLM in ``Q-fullTab'' while gains 14.76\%/8.98\%/13.88\% in ``Q-predCell'' setting, more than the best pipeline-based model TAPAS. For faithfulness metric PARENT and PARETN-T, \ourapproach provides the best performance among the pipeline models by outperforming TAPAS on the ``Q-predCell'' setting by 13.92\% and 3.88\% on PARENT and PARENT-T. Compared with end-to-end models, \ourapproach gives the best PARENT score while UniLM shows the best result regarding PARENT-T. It's explainable because \ourapproach incorporates information outside of the table to generate answers, achieving a trade-off between being grounded on the table and synthesizing informative answers. 

Furthermore, to answer Question 2 ``\textit{Are three stages of the framework necessary to generate high-quality answer?}'', we conduct an experiment in Table~
\ref{tab:res-main} by comparing the T5 model ``Q-fullTab'' with pipeline methods backend by T5 using ``Q-predCell''. The result shows proposed TAG for content selection TAGQA-T5 selecting 7\% of table cell outperforms T5 with fullTab. This indicates the table cell selection is necessary since relevant cells provide an anchor to generate high answer generation. Moreover, to investigate the effect of retrieval knowledge, we show results in Table~\ref{tab:res-main} by concatenating ``Retrieval'' to the input. The retrieval knowledge enhances model performance by providing background knowledge. The proposed model TAGQA-T5 provides the best result by integrating retrieval and informative selected cells. Lastly, our fusion module further enhanced the overall performance by aggregating tables and text efficiently.

Last but not least, to answer the question ``\textit{Is there space to further enhance performance using this framework ?}'', we conduct an oracle experiment shown in ``Oracle-T5''. With the simple Retrieval technique, T5 backend generation, and oracle table cell, the BLEU-4 result is 31\%, and PARENT, PARENT-T are over 30\%. If a better retrieval and fusion model is used, the model performance can be further boosted. 

\begin{table}[t]
\centering
\setlength{\tabcolsep}{3pt} 
\resizebox{0.98\textwidth/2}{!}{
\begin{tabular}{l|cc|cc}
\cline{2-2}
\toprule[1.5pt]
\textbf{Model} & \textbf{BLEU} & \textbf{METEOR} & ~~ \textbf{PARENT} & ~~ \textbf{PARENT-T}\\ \hline

\ourapproach      &  \textbf{31.84}   &   \textbf{30.16}   &  \textbf{29.59}  &   \textbf{28.26}  \\ \hline 
\ourapproach w/o JT &   31.35  &  29.65 & 28.93  & 27.48    \\ 
\ourapproach w/o SR   &  18.93  &  24.95    & 21.57    &  27.95 \\ 
\ourapproach w/o FiD   &  21.51   &   24.03  & 22.46   &  25.40 \\

\bottomrule[1.5pt]
\end{tabular}
}
\caption{Ablation study of the proposed model. We examine the ablated mode by removing the Joint Training~(JT) of \cs, Sparse Retrieval~(SR), and FiD.}
\label{tab:ablation}
\vspace{-1em}
\end{table}
\subsection{Analysis}
To further evaluate the quality of generated answer by various state-of-the-art models when compared to the ground-truth answer, we perform an additional human evaluation. Besides, we conduct an ablation study for \ourapproach to validate the three building blocks: jointly training of LM and GNN for \cs, external context retrieved from Wikipedia, and FiD model. Furthermore, a case study is presented which shows different answer qualities produced by various models.

\paragraph{Human Evaluation} 
Following \citep{fetaqa}, we recruit three human annotators who pass the College English Test (CET-6)\footnote{A national English as a foreign language test in China.} to judge the quality of the generated sentence. We randomly draw 100 samples from test examples in \fetaqa dataset and collect answers from \ourapproach and baseline models. Then, we present the generated answers to three human annotators without revealing the name of the model, thus reducing human variance.

We provide instructions for human raters to evaluate the sentence quality from four aspects: faithfulness, fluency, correctness, and adequateness. For each aspect, an annotator is supposed to assign a score ranging from 1 (worst) to 5 (best) based on the answer quality. The “overall” column refers to the average ranking
of the model. 
First, for fluency, the annotator checks if an answer is natural and grammatical. Second, for correctness, we compare the answer with the ground truth by checking if the predicted answer contains the correct information. Third, adequacy reflects if an answer contains all the aspects that are asked. Finally, faithfulness evaluates faithfulness if an answer is faithful and grounded to the contents of the highlighted table region such that it covers all the relevant information from the table while not including other key information outside of the table. From Table~\ref{tab:human-evaluation}, we can see \ourapproach ranked the top among all models.

\paragraph{Ablation Study}
To figure out which building blocks are driving the improvements, we examine different ablated models to understand each component of \ourapproach, including joint training of BERT and GNN from \cs, sparse retrieval, and FiD. Table~\ref{tab:ablation} presents the ablation results under different evaluation metrics. We can see that the model performance drops when any component is removed. Especially, ablating the sparse retrieval module results in the most drop in BLEU-4 and PARENT scores, while removing FiD causes the most significant drop in PARENT-T.

\begin{figure}[t]
    \centering
    \includegraphics[width=0.48\textwidth]{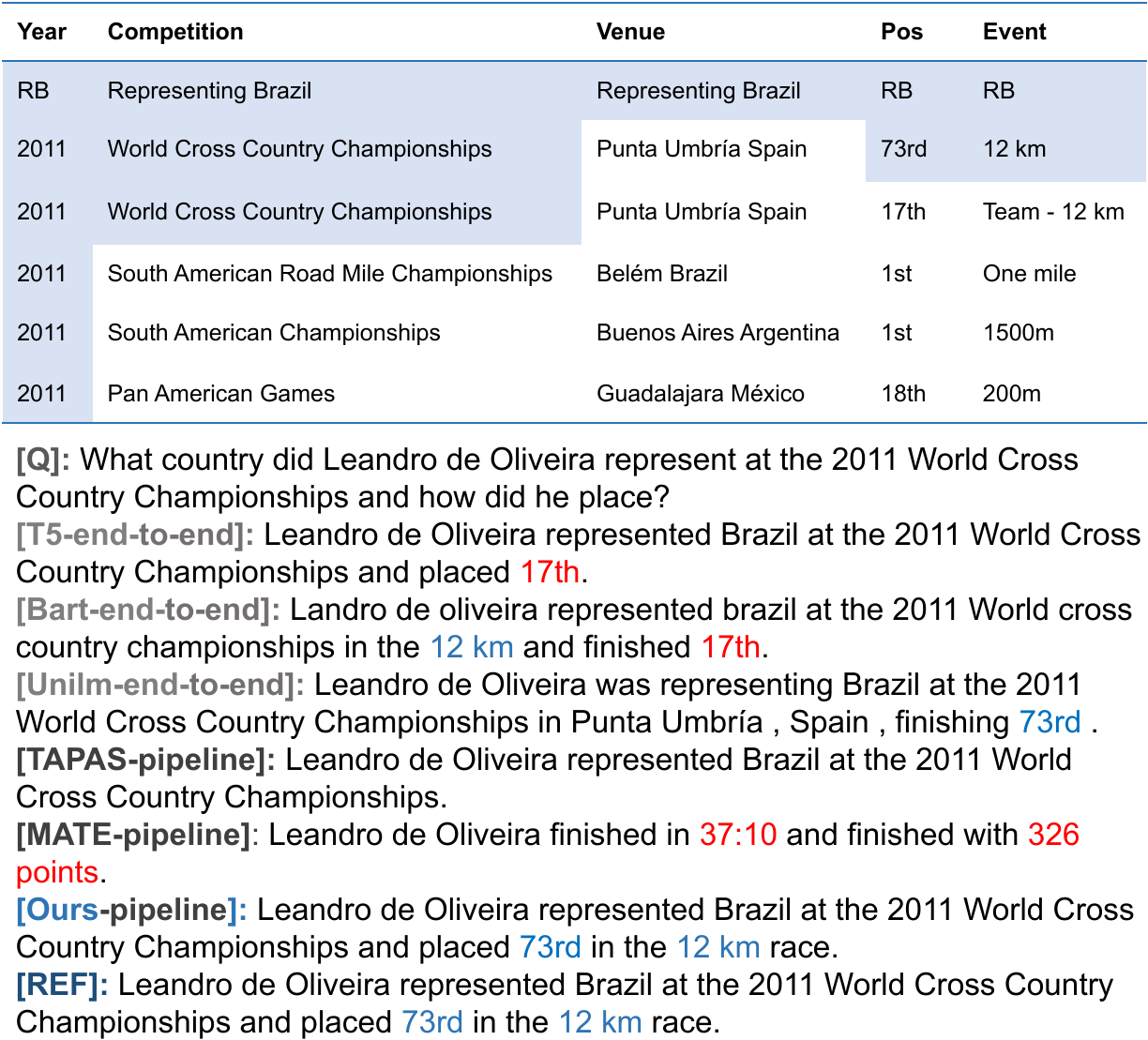}
    \caption{A case study from \fetaqa for qualitative analysis. The highlighted cells are the ground-truth relevant table cells. ``\textit{RB}'' refers to ``\textit{Representing Brazil}''. Hallucinated content from the predicted answer is marked in red and the correct content in blue.}
    \label{fig:case_study}
\end{figure}

\paragraph{Case Study} 
To inspect the effect of \ourapproach directly, we present a case study in Figure~\ref{fig:case_study}, where a sampled table, question, ground-truth relevant table cells (highlighted in blue), the predicted answers of models, as well as the reference are provided.
First, we find that the end-to-end model generally contains more information than pipeline models due to the more abundant table information while they suffer from hallucination. For example, T5 and BART identify the ranking position of ``\textit{Leandro de Oliveira}'' as ``\textit{17th}'' while it should be ``\textit{73rd}'' from the table. Second, for pipeline models, they tend to generate irrelevant information e.g. MATE mentions the duration and points instead of answering the ranking position and the event. Third, both the end-to-end and pipeline models (TAPAS) fail to cover all the relevant information from the table, e.g. UniLM did not capture the event 12km, and TAPAS fails to mention the position 73rd. By contrast, \ourapproach provides the highest table coverage while keeping the fluency of sentences.

\section{Related Work}

In this section, we review the related work to ours from the perspectives of TableQA, GNN for natural language processing, and knowledge-grounded text generation.

\paragraph{TableQA} \fetaqa is the first \tableqa dataset that addresses the significance of free-form answer generation, while most current research work including WikiTableQuestions~\citep{pasupat-liang-2015-compositional}, Spider~\citep{yu-etal-2018-spider}, HybridQA~\citep{chen2020hybridqa}, OTT-QA~\citep{chen2020open}, and TAT-QA~\citep{zhu-etal-2021-tat} focuses on the short factoid answer generation.
The early solution~\citep{zhong2017seq2sql,liang-etal-2017-neural} of addressing the \tableqa is to parse the natural question into a machine-executable
meaning representations that can be used to query the table. To reduce the labor-intensive logical annotation, a semantic parser trained over weak supervision from denotations has been drawing attention. 
Plenty of Transformer-based table pre-traininig models demonstrate decent \tableqa performance, e.g., TaPas~\citep{tapas}, MATE~\citep{mate}, TaBERT~\citep{yin20acl}, StruG~\citep{deng-etal-2021-structure}, GraPPa~\citep{yu2021grappa}, and TaPEx~\citep{liu2022tapex}.
In addition, rather than explore table structure,  RCI~\citep{glass-etal-2021-capturing} assumes the row and column are independent, and predicts the probability of containing the answer to a question in each row and column of a table individually.

\paragraph{GNN for Natural Language Processing} Apart from the extensively renowned causal language models that have showcased impressive results in various task~\citep{vaswani2017attention,parmar2018image,wang2023click,wang2022continuous,ner-wang-2023}, a rich variety of language processing tasks gain improvements from exploiting the power of GNN~\citep{li2015gated}. Tasks such as semantic parsing~\citep{chen-etal-2021-shadowgnn}, text classification~\citep{lin2021bertgcn}, text generation~\citep{fei-etal-2021-iterative}, question answering~\citep{wang2021gnn, yasunaga-etal-2021-qa} can be expressed with a graph structure and handled with graph-based methods. In addition, researchers apply GNN to model the text generation from structured data tasks e.g. graph-to-sequence~\citep{marcheggiani-perez-beltrachini-2018-deep}, and AMR-to-text~\citep{ribeiro-etal-2019-enhancing}.

\paragraph{Knowledge-Grounded Text Generation} Encoder-decoder-based models have been proposed to tackle the generation task by mapping the input to the output sequence. However, the input text is insufficient to provide knowledge to generate decent output due to the lack of commonsense, factual events, and semantic information. Knowledge-grounded text generation incorporating external knowledge such as linguistic features~\citep{liu-etal-2021-enriching}, knowledge graph~\citep{liu2021kg,li2021knowledge}, knowledge base~\citep{eric-manning-2017-copy,he-etal-2017-generating,liu-etal-2022-uni}, and textual knowledge~\citep{liu-dense-hierarchical,zhao-etal-2021-attend-memorize} help to generate a more logical and informative answer.

\section{Conclusion}
This paper presents a generalized pipeline-based framework \ourapproach for free-form long answer generation for \tableqa. The core idea of \ourapproach is to divide the answer generation process into three stages: (1) transform the table into a graph and jointly reason over the question-table graph to select relevant cells; (2) retrieve contextual knowledge from Wikipedia using sparse retrieval, and (3) integrate the selected cells with the content knowledge to predict the final answer. Extensive experiments on a public dataset \fetaqa are conducted to verify the generated answer quality from both the fluency and faithfulness aspects.

\section*{Limitations}
One limitation of \cs, which accepts the entire table as input, arises when dealing with large tables, as training both BERT and the graph model simultaneously becomes challenging due to GPU memory constraints.
Consequently, one promising avenue for future research involves the efficient modeling of large tables.
Furthermore, it's worth noting that the availability of only one public dataset, \fetaqa, for free-form \tableqa, has constrained our validation efforts to this single dataset. However, we are committed to expanding the scope of our research in the future by evaluating the performance of our pipeline model, \ourapproach, across multiple free-form \tableqa datasets.


\bibliography{anthology}
\bibliographystyle{acl_natbib}

\end{document}